\documentclass[11pt,conference]{IEEEtran}
\IEEEoverridecommandlockouts
\usepackage[letterpaper,top=0.8in,bottom=0.8in,left=0.625in,right=0.625in]{geometry}
\usepackage{cite}
\usepackage{amsmath,amssymb,amsfonts}
\usepackage{graphicx}
\usepackage{textcomp}
\usepackage{booktabs}
\usepackage{multirow}
\usepackage{url}
\usepackage{tikz}

\newcommand\copyrighttext{%
  \footnotesize \textcopyright~2026 IEEE. Personal use of this material is
  permitted. Permission from IEEE must be obtained for all other uses, in
  any current or future media, including reprinting/republishing this
  material for advertising or promotional purposes, creating new collective
  works, for resale or redistribution to servers or lists, or reuse of any
  copyrighted component of this work in other works. Accepted for
  publication at the 2026 IEEE 17th Annual Ubiquitous Computing,
  Electronics \& Mobile Communication Conference (UEMCON).}
\newcommand\copyrightnotice{%
\begin{tikzpicture}[remember picture,overlay]
\node[anchor=south,yshift=8pt] at (current page.south)
  {\parbox{\dimexpr\textwidth-2\fboxsep\relax}{\copyrighttext}};
\end{tikzpicture}%
}

\def\BibTeX{{\rm B\kern-.05em{\sc i\kern-.025em b}\kern-.08em
    T\kern-.1667em\lower.7ex\hbox{E}\kern-.125emX}}

\newcommand{\method}{BudgetMem}

\begin{document}

\title{BudgetMem: Training-Free Selective Memory for Cost-Efficient Long-Context Processing in Language Models}

\author{\IEEEauthorblockN{Chandra Vamsi Krishna Alla}
\IEEEauthorblockA{\textit{Dept. of Computer Science and Engineering} \\
\textit{University of Texas at Arlington}\\
Arlington, TX, USA \\
cka0054@mavs.uta.edu}
\and
\IEEEauthorblockN{Harish Naidu Gaddam}
\IEEEauthorblockA{\textit{Dept. of Computer Science and Engineering} \\
\textit{University of Texas at Arlington}\\
Arlington, TX, USA \\
hxg6016@mavs.uta.edu}
\and
\IEEEauthorblockN{Manohar Kommi}
\IEEEauthorblockA{\textit{Dept. of Computer Science and Engineering} \\
\textit{University of Texas at Arlington}\\
Arlington, TX, USA \\
mxk8106@mavs.uta.edu}
\and
\IEEEauthorblockN{Sheikh Nazib Ahmed}
\IEEEauthorblockA{\textit{Dept. of Information Systems and Operations Management (ISOM)} \\
\textit{University of Texas at Arlington}\\
Arlington, TX, USA \\
sxa5256@mavs.uta.edu}
}

\maketitle
\copyrightnotice
\begin{abstract}
Processing long documents with large language models (LLMs) is expensive: a single query over a 100K-token document can cost from tens of cents to over a dollar in API fees, depending on the model, and memory grows linearly with context length. We introduce \method{}, a training-free architecture that keeps only high-salience content under an explicit memory budget. Unlike token-level neural compressors such as LLMLingua, \method{} makes chunk-level keep-or-discard decisions from interpretable features: entity density, TF-IDF importance, position, numerical density, discourse markers, and question presence. Across four benchmarks, \method{} matches the uncompressed baseline on template-generated structured documents (F1 = 0.859 vs.\ 0.855) while discarding 70\% of chunks. As a pre-retrieval filter on the same documents it outperforms LLMLingua-2 by a wide margin (0.859 vs.\ 0.554), because token-level compression destroys the phrasal structure that retrieval depends on; the advantage holds across model scale and family (Llama-3.2-3B, Qwen2.5-7B), under dense retrieval (0.918 vs.\ 0.530), and in LLMLingua-2's intended direct-feed mode (0.797 vs.\ 0.720). On real NLP research papers (Qasper), \method{} is competitive but not free, trailing the full baseline by 7.4\% and LLMLingua-2 by 5.5\% relative F1 at a 72\% storage reduction. Removing discourse markers or numerical density hurts most, although TF-IDF alone scores as well on template documents. The pipeline needs no trained models or GPU for compression and runs on a \$10/month Colab instance.
\end{abstract}


\section{Introduction}
\label{sec:introduction}

The context windows of modern large language models have expanded dramatically. GPT-4~\cite{openai2023gpt4} supports 128K tokens, Claude~\cite{anthropic2024claude} accepts up to 200K tokens, and recent open-source models like Llama~3~\cite{dubey2024llama3} handle 128K tokens natively.

Yet using these long contexts remains expensive: memory and latency grow with sequence length, and a legal firm analyzing thousands of 50K--100K-token contracts pays tens of cents to over a dollar per document in API fees, depending on the model.

Existing remedies either modify the architecture, which still requires the full input in memory at inference, or compress the context with a second neural model (Section~\ref{sec:related}). We take a different path. Instead of compressing text at the token level using learned models, \method{} operates at the \emph{chunk level}: it splits a document into paragraph-sized pieces, scores each with simple interpretable features, and permanently discards chunks below a budget threshold. The surviving chunks are indexed with BM25. No neural models are involved in the compression step---just entity counts, TF-IDF statistics, position heuristics, and discourse markers, all running on CPU in well under a second, even for a 50K-token document.
The motivation is a linguistic observation: a complete paragraph carries far more recoverable meaning than a scattered set of individually high-importance words plucked from different sentences. Token-level pruning destroys syntactic structure and co-reference chains, producing fragments like ``John traveled\ldots Paris saw\ldots historical museums'' that defeat both human comprehension and automated retrieval. Chunk-level selection preserves complete thoughts.

Our experiments support this intuition. On template-generated structured documents, \method{} matches the uncompressed baseline while discarding 70\% of chunks, whereas LLMLingua-2 used as the same pre-retrieval filter loses 35\% F1. On real Qasper papers the picture is more mixed (Section~\ref{sec:results}).

Our contributions: (1)~a \emph{training-free selective memory architecture} that matches baseline RAG on template-generated structured documents while discarding 70\% of chunks, using only interpretable features and no GPU for compression; (2)~a \emph{controlled granularity study} (token vs.\ chunk, statistical vs.\ neural scoring) with per-feature ablation; (3)~\emph{evaluation across four benchmarks}, three pipeline modes, and two model families; and (4)~a \emph{practical deployment} on a \$10/month Colab instance.

\section{Related Work}
\label{sec:related}

\textbf{Long-context transformers.} The standard Transformer~\cite{vaswani2017attention} has $O(n^2)$ attention cost. Sparse patterns~\cite{child2019generating,zaheer2020bigbird,beltagy2020longformer}, linear attention~\cite{choromanski2021rethinking,katharopoulos2020transformers}, recurrence and memory compression~\cite{dai2019transformerxl,rae2019compressive}, and position-encoding extensions~\cite{su2021roformer,press2022train} cut computation or extend length, but still hold the full input in memory. \method{} shrinks the input \emph{before} it reaches the model and complements all of them.

\textbf{Context compression.} LLMLingua~\cite{jiang2023llmlingua} prunes tokens by small-LM perplexity; LLMLingua-2~\cite{pan2024llmlingua2} casts compression as BERT-based token classification; Selective Context~\cite{li2023compressing} uses self-information; RECOMP~\cite{xu2023recomp} and AutoCompressor~\cite{chevalier2023adapting} train compressors. All need neural inference to compress. \method{} works at the chunk level, preserving syntactic coherence, with no neural model in the compression step.

\textbf{Retrieval and memory.} Standard RAG~\cite{lewis2020retrieval,karpukhin2020dense,izacard2021leveraging,shi2023replug,borgeaud2022improving} indexes all content and filters at query time. \method{} filters at ingestion instead, so low-salience material never enters the index, and it can sit in front of any RAG system. Memory-augmented LLMs~\cite{wang2024memlong,xiao2024infllm,wang2024memorag,zhang2025amem} maintain dynamic memory during generation; our static, budget-aware storage decisions are orthogonal.

\section{Method}
\label{sec:method}

\method{} takes a long document $D$ and produces a compact, retrievable memory store $\mathcal{S}$ through four stages: chunking, salience scoring, budget-aware selection, and BM25 indexing. At query time, the system retrieves relevant chunks from $\mathcal{S}$ and generates an answer using a base LLM. All compression decisions are made at ingestion time using interpretable, non-neural features.

\subsection{Document Chunking}

Given a document $D$ with $N$ tokens, we split it into $M$ overlapping chunks $\{C_1, \ldots, C_M\}$ of size $c = 150$ with overlap $o = 30$. The overlap keeps information that spans chunk boundaries.

\subsection{Salience Scoring}

Each chunk $C_i$ receives a composite salience score:
\begin{equation}
    s_i = \sum_{j=1}^{6} w_j \cdot f_j(C_i)
\end{equation}
where $w_j$ is the weight for feature $j$ and $f_j(C_i)$ is the normalized feature value, computed from six interpretable features:

\textbf{Entity density} ($w_1 = 0.20$): The ratio of capitalized tokens to total tokens, a lightweight proxy for named entity density. Capitalized words correlate with proper nouns---names, organizations, locations---signaling factual, query-relevant content. On academic text the proxy is noisy (citation-heavy and acknowledgment passages saturate it), which is why it contributes little in the ablation of Section~\ref{sec:analysis}.

\textbf{TF-IDF importance} ($w_2 = 0.20$): The mean TF-IDF score of the chunk's tokens relative to the document-level corpus, capturing chunks with distinctive vocabulary.

\textbf{Position bias} ($w_3 = 0.15$): A tent-shaped prior that is highest for chunks in the middle of the document and decays linearly to 0.3 at either end:
\begin{equation}
    f_3(C_i) = \min\!\left(1.0,\; \max\!\left(0,\; 1.3 - 2.0 \cdot \left| \tfrac{i}{M} - 0.5 \right|\right)\right)
\end{equation}
In a typical paper the middle holds the method and results sections that factual questions target, while the ends hold boilerplate (title block, acknowledgments, references). The ablation in Section~\ref{sec:analysis} shows this prior is nearly inert; it is best read as a mild tie-breaker.

\textbf{Numerical density} ($w_4 = 0.15$): The proportion of tokens that are numeric (integers, decimals, dates, percentages). Passages containing statistics, measurements, and quantities are frequently the targets of factual questions.

\textbf{Discourse markers} ($w_5 = 0.10$): The density of discourse connectives, detected by case-insensitive substring matching against a fixed 48-entry lexicon of contrastive, causal, additive, exemplifying, and sequencing connectives (e.g., ``however,'' ``therefore,'' ``for example,'' ``in contrast,'' ``finally''), counted per chunk and divided by chunk length. The lexicon is generic---no domain terms---and substring matching is a knowing simplification (``after'' also fires inside ``thereafter''). These markers signal argumentative structure and topic transitions.

\textbf{Question presence} ($w_6 = 0.10$): Whether the chunk contains interrogative sentences, detected via question marks or interrogative words (who, what, where, when, why, how).

All features are normalized to $[0, 1]$ across the document's chunks. Weights were set once on a held-out set of 20 documents and \emph{held fixed across all benchmarks}. Because the features are interpretable, practitioners can see exactly why any chunk was kept or discarded, which matters in legal, medical, and regulatory settings.

\subsection{Budget-Aware Selection and Retrieval}

Given a retention ratio $r$, we select the top $\lfloor r \cdot M \rfloor$ chunks by salience. The remaining chunks are permanently discarded. Our default is $r = 0.30$, yielding 70\% storage savings. Retained chunks are indexed with BM25. At query time, we retrieve the top $k = 3$ chunks and provide them as context to the base LLM. We use Llama-3.2-3B-Instruct~\cite{dubey2024llama3} in FP16 precision with greedy decoding. Chunking, scoring, and selection run once per document at ingestion; only retrieval and generation run per query.

\section{Experimental Setup}
\label{sec:experiments}

\textbf{Datasets.} We evaluate across four benchmarks. \emph{SQuAD v2.0}~\cite{rajpurkar2018know}: 500 Wikipedia articles averaging 237 tokens (range 148--448), testing the short-document regime. \emph{Structured academic papers}: 200 QA pairs on 40 template-generated research papers of about 3,700 tokens each, a controlled benchmark with clear section structure in which each section repeats one templated paragraph 5--10 times, so answer-bearing sentences appear many times. \emph{Qasper}~\cite{dasigi2021qasper}: 102 QA pairs on real NLP research papers averaging 3,608 tokens, validating \method{} on naturally occurring academic text. \emph{NarrativeQA}~\cite{kocisky2018narrativeqa}: 50 book excerpts and movie scripts of 50K--100K tokens---unstructured narrative prose that pushes all methods to their limits.
\textbf{Infrastructure.} All QA experiments use Llama-3.2-3B-Instruct in FP16 with greedy decoding (max 100 new tokens) on one NVIDIA T4 GPU (15GB) in a \$10/month Google Colab Pro instance; the cross-model check uses Qwen2.5-7B-Instruct in 4-bit NF4 on the same GPU. Code, notebooks, and result files are available at \url{https://github.com/Chandra-Alla/BudgetMem-paper}.

\textbf{Baselines.} (1)~\emph{Baseline RAG}: all chunks retained, BM25 retrieval. (2)~\emph{Token-level TF-IDF}: top 30\% of tokens by TF-IDF. (3)~\emph{LLMLingua-2}~\cite{pan2024llmlingua2}: a BERT-based token classifier run at rate 0.3 (nominal 30\% retention). (4)~\emph{Naive chunk strategies}: Random, First-N, Last-N, TF-IDF-only, each at 30\% budget. On medium documents we additionally vary the pipeline itself: (5)~\emph{dense retrieval}, cosine top-3 over all-MiniLM-L6-v2 embeddings in place of BM25; and (6)~\emph{direct feed}, the compressor's entire output placed in the context window with no retrieval stage---LLMLingua-2's intended mode. NarrativeQA experiments use 3 random seeds (42, 123, 456); the other benchmarks use a single greedy-decoding run.

\textbf{Metric.} Token-level F1 between predicted and ground-truth answers after standard normalization (lowercasing, article removal, punctuation stripping).

\section{Results}
\label{sec:results}

\subsection{Short Documents (SQuAD)}

On 500 short Wikipedia articles, \method{} trades 9.7\% F1 (0.723 vs.\ 0.801 for baseline RAG) for only 15.5\% storage savings, while adding 16\% latency (0.43s vs.\ 0.37s). With few chunks per document, most score highly and little low-value material can be safely discarded---this negative finding is instructive: we recommend full-context processing below $\sim$1K tokens.

\subsection{Medium Documents (Structured Papers)}

This is \method{}'s strongest regime (Table~\ref{tab:medium_results}): on 200 QA pairs over template-generated papers of about 3,700 tokens, it \emph{matches} the uncompressed baseline (F1 = 0.859 vs.\ 0.855) while retaining only 30\% of chunks. All medium-document configurations (Tables~\ref{tab:medium_results}, \ref{tab:budget_sensitivity}, and~\ref{tab:naive_comparison}) were run in one session through one code path, so their numbers are directly comparable. LLMLingua-2's compressor is not bitwise deterministic on GPU; an earlier run scored 0.532.

\begin{table}[t]
\centering
\caption{Medium documents (200 QA pairs, 30\% budget): F1 under sparse (BM25) and dense (MiniLM) retrieval over the stored content, direct-feed with no retrieval (the baseline's direct column is the full uncompressed document in-context), and the cross-model check. Storage is the measured fraction kept: 24\% of words for LLMLingua-2 (rate 0.3), 29\% of chunks for \method{}.}
\label{tab:medium_results}\label{tab:cross_model}
\resizebox{\columnwidth}{!}{%
\begin{tabular}{lccccc}
\toprule
 & & \multicolumn{3}{c}{\textbf{Llama-3.2-3B}} & \textbf{Qwen2.5-7B} \\
\cmidrule(lr){3-5}\cmidrule(lr){6-6}
\textbf{Method} & \textbf{Storage} & \textbf{BM25} & \textbf{Dense} & \textbf{Direct} & \textbf{BM25} \\
\midrule
Baseline RAG & 100\% & 0.855 & 0.844 & 0.877 & 0.900 \\
LLMLingua-2 & 24\% & 0.554 & 0.530 & 0.720 & 0.773 \\
\textbf{BudgetMem (Ours)} & \textbf{29\%} & \textbf{0.859} & \textbf{0.918} & \textbf{0.797} & \textbf{0.960} \\
\bottomrule
\end{tabular}}
\end{table}

\textbf{Sparse retrieval (BM25).} The most striking result is LLMLingua-2's 35\% relative F1 degradation (0.554). Token-level compression destroys the phrasal structure BM25 relies on: compressed text like ``novel approach learning architecture evaluate accuracy'' no longer contains the multi-word spans needed for effective sparse retrieval. \method{}'s chunk-level preservation avoids this failure mode.

\textbf{Dense retrieval.} A natural objection is that this advantage is a BM25 artifact. It is not: with cosine retrieval over all-MiniLM-L6-v2 embeddings, LLMLingua-2 scores 0.530 while \method{} \emph{rises} to 0.918, above the dense-retrieval baseline (0.844). Fragmented token-level output embeds as poorly as it BM25-retrieves, whereas salience filtering appears to help the dense retriever by removing distractor chunks before they can be embedded.

\textbf{Direct feed.} LLMLingua-2 was designed to place its output directly in the LLM context, not behind a retriever. In that mode (whole compressed document in context, limits raised so nothing is truncated) it recovers to 0.720, but \method{}'s retained chunks fed the same way score 0.797; both trail the full document in context (0.877). So the pre-retrieval result does not merely catch LLMLingua-2 out of position: at a similar nominal budget, chunk-level selection also leads in LLMLingua-2's own operating mode ($+$0.077, 95\% CI $[+0.012, +0.142]$, $p = 0.02$), with no GPU for compression, though the margin is modest and LLMLingua-2 kept somewhat less text (Table~\ref{tab:medium_results}).

\subsection{Cross-Model Validation (Qwen2.5-7B)}

A second concern is whether the advantage is specific to one base model. Replicating the BM25 comparison on Qwen2.5-7B-Instruct---a different parameter scale and model family, under 4-bit (NF4) quantization on the same T4---preserves the ranking (last column of Table~\ref{tab:medium_results}): \method{} reaches 0.960 vs.\ LLMLingua-2's 0.773, matching or exceeding the uncompressed baseline (0.900). The larger model narrows LLMLingua-2's deficit (35\% relative at 3B, 14\% at 7B) because it reconstructs meaning from fragmented input better; chunk-level selection stays ahead even so.

\subsection{Very Long Documents (NarrativeQA)}

Table~\ref{tab:narrativeqa_results} compares methods at a nominal 30\% budget.

\begin{table}[t]
\centering
\caption{Real and very long documents (nominal 30\% budget): Qasper (real NLP research papers, 102 QA pairs, 3.6K tokens) and NarrativeQA (50 examples, 50K--100K tokens, mean $\pm$ std over 3 seeds).}
\label{tab:narrativeqa_results}\label{tab:qasper_results}
\resizebox{\columnwidth}{!}{%
\begin{tabular}{lcccc}
\toprule
\textbf{Method} & \textbf{Storage} & \textbf{Neural?} & \textbf{Qasper F1} & \textbf{NarrativeQA F1} \\
\midrule
Baseline RAG & 100\% & -- & 0.179 & $0.0471 \pm 0.0090$ \\
Token TF-IDF & 30\% & No & -- & $0.0007 \pm 0.0000$ \\
LLMLingua-2 & 30\% & Yes & 0.175 & $0.0402 \pm 0.0026$ \\
\textbf{BudgetMem (Ours)} & \textbf{30\%} & \textbf{No} & \textbf{0.165} & $\mathbf{0.0351 \pm 0.0022}$ \\
\bottomrule
\end{tabular}}
\end{table}

Simple token-level TF-IDF collapses to near-zero F1 (0.0007): removing scattered tokens from massive texts destroys them beyond retrieval. Neural token scoring (0.0402) and simple chunk-level selection (0.0351) both avoid this collapse, but \method{} trails the uncompressed baseline (0.0471) by 25\% relative. Scores sit near the floor for every method at 3B, and the gap between LLMLingua-2 and \method{} is smaller than the baseline's spread across seeds, so we read this benchmark only as evidence that chunk-level selection does not collapse, not as a ranking.

\subsection{Real Academic Papers (Qasper)}

To validate \method{} beyond controlled documents, we evaluate on 102 QA pairs from Qasper~\cite{dasigi2021qasper} (Table~\ref{tab:qasper_results}). Documents average 3,608 tokens, about the same length as our template papers.

On Qasper, \method{} trails both the uncompressed baseline (by 7.4\% relative F1) and LLMLingua-2 (by 5.5\%; 0.165 vs.\ 0.175), for a measured 72\% storage saving. With 102 questions these gaps are within noise (unpaired Welch $p > 0.6$), so Qasper shows competitive rather than matched performance. Length does not explain the difference from the template results: Qasper and template papers both have $\approx$30 chunks, of which a 30\% budget keeps $\approx$9. Two other differences likely do. Template papers repeat each section's paragraph 5--10 times, so discarding 70\% of chunks still leaves a copy of nearly every answer, whereas a real paper states each fact once and any discarded chunk may hold the answer; and the feature weights were set on template documents whose answer-bearing passages are unusually number-dense. Absolute F1 is low for every system because Qasper answers are free-form and a 3B model without domain adaptation rarely reproduces them verbatim. The conclusion is a scoped one: on real papers, \method{} buys a 72\% storage reduction at a point-estimate cost of 7.4\% relative F1, rather than matching the baseline as it does on redundant template documents.

\subsection{Benchmark Comparison and Budget Sensitivity}

\begin{figure}[t]
\centering
\includegraphics[width=\columnwidth]{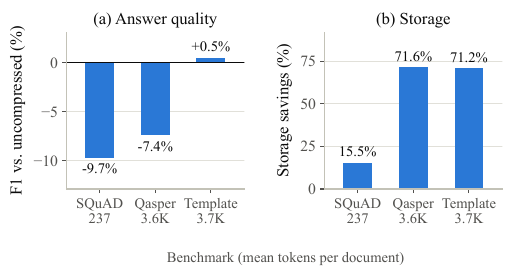}
\caption{Cost and savings by benchmark at the 30\% budget. (a)~F1 relative to the uncompressed baseline; (b)~measured storage savings. Qasper and the template papers have similar lengths, so the gap between them likely reflects differences between the corpora, such as redundancy and answer type, rather than length. NarrativeQA is omitted because every method sits near the F1 floor there (Table~\ref{tab:narrativeqa_results}).}
\label{fig:length_scaling}
\end{figure}

Fig.~\ref{fig:length_scaling} summarizes cost and savings by benchmark. Short documents gain little: on 237-token SQuAD passages \method{} saves only 15.5\% of storage for a 9.7\% F1 cost, because there is little low-value material to discard. At similar lengths ($\sim$3.6--3.7K tokens), real Qasper papers cost 7.4\% while redundant template papers cost nothing, both at 71--72\% savings, which points to differences between the corpora, such as redundancy, rather than length. On NarrativeQA every method sits near the F1 floor. Table~\ref{tab:budget_sensitivity} sweeps 7 retention ratios on all 200 pairs: 20--30\% retention is the sweet spot, saving 71--81\% of storage while matching or exceeding the 0.855 baseline. The sweep is not strictly monotone (the 70\% point dips to 0.795, likely an artifact of top-3 retrieval interacting with which chunks survive at each threshold); robustly, 20--30\% recovers baseline quality while 10\% degrades sharply.

\begin{table}[t]
\centering
\caption{Budget sensitivity on structured documents (200 QA pairs; baseline F1 0.855).}
\label{tab:budget_sensitivity}
\resizebox{\columnwidth}{!}{%
\begin{tabular}{lccccccc}
\toprule
\textbf{Budget} & 10\% & \textbf{20\%} & \textbf{30\%} & 40\% & 50\% & 70\% & 90\% \\
\midrule
F1 & 0.689 & \textbf{0.868} & \textbf{0.859} & 0.856 & 0.848 & 0.795 & 0.862 \\
Savings & 90.4\% & \textbf{80.8\%} & \textbf{71.2\%} & 61.7\% & 51.1\% & 32.0\% & 12.8\% \\
\bottomrule
\end{tabular}}
\end{table}

\section{Analysis}
\label{sec:analysis}

\subsection{Naive Baseline Ablation}

Table~\ref{tab:naive_comparison} compares \method{} against four naive selection strategies at the same 30\% budget, run in the same session and on the same 200 pairs as Table~\ref{tab:medium_results}. Three observations. First, \emph{granularity dominates scoring}: even random chunk selection (0.785) far exceeds token-level LLMLingua-2 (0.554)---retaining coherent chunks matters more than how they are chosen. Second, positional heuristics fail: keeping the first 30\% of chunks scores 0.357, because answers to factual questions concentrate mid-document---exactly what the position prior of Section~\ref{sec:method} encodes. Third, chunk-level TF-IDF alone---itself an instance of the \method{} framework with a single feature---edges out the full six-feature scorer here (0.880 vs.\ 0.859). Together with the ablation below, this locates the framework's value in \emph{budgeted chunk-level selection and interpretability}, not in any particular weight vector; on template documents, one distinctive-vocabulary signal is already near-optimal. We have not re-run the naive selectors on the real-paper benchmarks, so we make no cross-corpus claim about the best feature mix.

\begin{table}[t]
\centering
\caption{Chunk-selection variants on medium documents (30\% budget, 200 QA pairs, same run as Table~\ref{tab:medium_results}). Gap is relative to full \method{}.}
\label{tab:naive_comparison}\label{tab:feature_ablation}
\begin{tabular}{lcc}
\toprule
\textbf{Configuration} & \textbf{F1 Score} & \textbf{Gap} \\
\midrule
\textbf{BudgetMem (all 6 features)} & \textbf{0.859} & -- \\
\midrule
\multicolumn{3}{l}{\textit{Naive selection}} \\
Random 30\% & 0.785 & $-$8.6\% \\
First 30\% & 0.357 & $-$58.5\% \\
Last 30\% & 0.527 & $-$38.7\% \\
TF-IDF only & 0.880 & $+$2.4\% \\
\midrule
\multicolumn{3}{l}{\textit{Leave-one-out feature ablation}} \\
w/o Discourse markers & 0.605 & $-$29.6\% \\
w/o Numerical density & 0.718 & $-$16.4\% \\
w/o TF-IDF importance & 0.853 & $-$0.8\% \\
w/o Position bias & 0.862 & $+$0.3\% \\
w/o Entity density & 0.859 & 0.0\% \\
w/o Question presence & 0.859 & 0.0\% \\
\bottomrule
\end{tabular}
\end{table}

\subsection{Per-Feature Ablation}

To quantify each feature's contribution, we run a leave-one-out ablation on the same 200 QA pairs (lower half of Table~\ref{tab:feature_ablation}).

Removing \textbf{discourse markers} ($-$29.6\%) or \textbf{numerical density} ($-$16.4\%) hurts most: these features separate substantive, number-bearing passages from filler, and without either one the noisier features likely dominate the score. That TF-IDF alone scores 0.880 (upper half of Table~\ref{tab:naive_comparison}) shows the other features are not needed on these documents; what matters is that noisy signals do not outvote informative ones. Entity density and question presence show zero impact here: papers contain few embedded questions, and their capitalization mostly marks citations and boilerplate rather than answers.

\textbf{Weight sensitivity.} The hand-tuned weights are not a knife-edge. We perturbed each weight by $\pm$10\% and $\pm$20\% (24 configurations) and drew 20 random joint perturbations with every weight scaled by $U(0.8, 1.2)$. Across all 44 configurations, the retained-chunk sets overlapped the default by at least 90\% (Jaccard; mean above 99\%), and F1 stayed within $[0.858, 0.864]$ of the default 0.859 ($-$0.2\% to $+$0.6\%). Behaviour is governed by which features are present (the ablation above), not by their precise weights.

\textbf{Statistical significance.} On the 200 medium-document pairs, a conservative unpaired Welch test puts the \method{}--baseline difference at $+$0.004 (95\% CI $[-0.039, +0.047]$, $p = 0.85$): the two are statistically indistinguishable, which supports our ``matches the baseline'' claim rather than a claim of improvement. The margin over LLMLingua-2 is unambiguous---$+$0.306 under BM25 (95\% CI $[+0.237, +0.375]$, $p < 10^{-17}$) and $+$0.389 under dense retrieval (95\% CI $[+0.324, +0.453]$, $p < 10^{-31}$)---as is \method{}'s $+$0.074 margin over the uncompressed baseline under dense retrieval (95\% CI $[+0.033, +0.116]$, $p < 10^{-3}$). Because all systems answer the same questions, a paired analysis would give tighter intervals; these are conservative.

\textbf{What gets kept.} Real Qasper papers make the behaviour concrete. In a paper on low-resource machine translation, the highest-salience chunk is the experimental setup (corpus and dev-set sizes, toolkit, hyper-parameters; numerical density 1.0, entity density 0.83) and the lowest is the one-paragraph conclusion (numerical density 0)---substantive prose that surface features cannot tell from filler. Acknowledgments show the opposite: every word capitalized, entity density saturated at 1.0, yet low TF-IDF and zero numerical density still discard them. Both are genre effects: the features reward the number- and name-dense passages that factual questions about papers target; narrative text, conversational logs, or ``why'' questions would need re-weighting or a domain lexicon---with an interpretable scorer, a matter of editing a list rather than retraining a model.

\subsection{Computational Cost}

\method{}'s compression (chunking, feature scoring with scikit-learn's TF-IDF vectorizer, and selection) needs only a CPU: on a laptop CPU it takes 0.05s for a 52K-token document and 0.004s for a 3.7K-token template paper (median of 7 runs), with no GPU memory. LLMLingua-2 instead runs a BERT-base classifier over every token of the document, which in practice calls for a GPU; we did not benchmark its latency under matched conditions.

Per-query prompt size is set by top-3 retrieval ($\sim$450 tokens) with or without \method{}; its savings are in index storage and in the indexing or embedding work done at ingestion, which shrink with the 70\% of chunks discarded.

\subsection{When to Use \method{}}
\label{sec:when}

\textbf{Use it} for sectioned documents with substantial redundancy, like the templated papers in our benchmark, where a 20--30\% retention budget saved 71--81\% of index storage at baseline-level F1 (Table~\ref{tab:budget_sensitivity}); when compression must run on CPU; when decisions must be auditable (every retained chunk carries an inspectable feature breakdown); or when the downstream retriever is sparse (BM25), which token-level compression breaks. \textbf{Expect a cost} on ordinary papers that state each fact once: 7.4\% relative F1 on real NLP papers of $\sim$3.6K tokens (Qasper). \textbf{Do not use it} below $\sim$1K tokens, where there is little to discard (SQuAD: $-$9.7\% F1 for 15.5\% savings), or on unstructured narrative prose, where it trails the baseline by 25\% relative (NarrativeQA). \textbf{Untested:} conversational logs, multilingual text, summarization, and multi-hop reasoning; the discourse lexicon and numerical-density feature are tuned to English expository text. \textbf{Consider LLMLingua-2} when a GPU is available and documents are not redundant; on Qasper it scored slightly higher, though within noise.

\section{Conclusion}
\label{sec:conclusion}

We presented \method{}, a training-free selective memory system that scores document chunks with interpretable features and discards low-salience material at ingestion time. It matches baseline quality on template-generated structured documents while cutting index storage by 70\%, and on those documents outperforms neural token-level compression as a pre-retrieval filter, without trained models, GPU-based compression, or expensive hardware.

Three findings emerge. First, granularity matters: token-level compression badly degrades retrieval-based QA (LLMLingua-2 loses 35\% F1 on template documents), while chunk-level selection stays ahead of LLMLingua-2 under sparse retrieval, dense retrieval, and direct feed. Second, simple features carry enough signal on structured text; removing discourse markers or numerical density hurts most, yet TF-IDF alone does as well. Third, on real papers (Qasper) \method{} trades a 7.4\% relative F1 cost for 72\% storage savings, which defines an operating envelope (Section~\ref{sec:when}) rather than a universal win.

\textbf{Limitations.} Our structured benchmark is template-generated: each section repeats one paragraph several times and the answers are numeric hyper-parameters, which makes aggressive filtering nearly lossless, favors numerical density, and inflates absolute F1 relative to real papers; the dense-retrieval, direct-feed, and cross-model checks use it too, so the Qasper results are the more conservative estimate of \method{}'s value. Evaluation covers only extractive QA in English; summarization, multi-hop reasoning, and multilingual settings remain untested, as does a full multi-benchmark sweep beyond the 7B cross-model check. The dense-retrieval pilot uses one small encoder (all-MiniLM-L6-v2); stronger encoders may shift the picture. Weights are manually tuned---learned policies would likely help, at the cost of the training-free property.

\textbf{Future work.} Learned write policies, adaptive per-document budgets, and token-level refinement within retained chunks could close the remaining gaps while staying cheaper than neural compression.

\bibliographystyle{IEEEtran}
\bibliography{references}

\end{document}